\documentclass[fullpaper,cameraready]{nldl}

\renewcommand{\DOI}{10.7557/18.6269}

\usepackage[utf8]{inputenc}
\usepackage{url}
\usepackage{graphicx}
\usepackage{authblk}

\usepackage{multirow}
\usepackage{booktabs}
\usepackage{adjustbox}
\usepackage{svg}

\usepackage{algorithm}
\usepackage{algorithmic}

\usepackage[square,numbers]{natbib}
\usepackage{hyperref}

\title{An analysis of over-sampling labeled data in semi-supervised learning with FixMatch}

\author[1,2]{Miquel Martí i Rabadán\thanks{Corresponding Author: miquelmr@kth.se}}
\author[1]{Sebastian Bujwid}
\author[2]{Alessandro Pieropan}
\author[1]{Hossein Azizpour}
\author[1]{Atsuto Maki}

\affil[1]{KTH Royal Institute of Technology, Sweden}
\affil[2]{Univrses AB, Sweden}

\date{\vspace{-5ex}}

\begin{document}
\nldlmaketitle

\begin{abstract}

Most semi-supervised learning methods over-sample labeled data when constructing training mini-batches. This paper studies whether this common practice improves learning and how. We compare it to an alternative setting where each mini-batch is uniformly sampled from all the training data, labeled or not, which greatly reduces direct supervision from true labels in typical low-label regimes.
However, this simpler setting can also be seen as more general and even necessary in multi-task problems where over-sampling labeled data would become intractable.
Our experiments on semi-supervised CIFAR-10 image classification using FixMatch show a performance drop when using the uniform sampling approach which diminishes when the amount of labeled data or the training time increases. 
Further, we analyse the training dynamics to understand how over-sampling of labeled data compares to uniform sampling. Our main finding is that over-sampling is especially beneficial early in training but gets less important in the later stages when more pseudo-labels become correct. Nevertheless, we also find that keeping \textit{some} true labels remains important to avoid the accumulation of confirmation errors from incorrect pseudo-labels.

\end{abstract}

\section{Introduction}
\label{sec:intro}
Semi-supervised learning has recently established itself as an effective approach to greatly reduce the annotation costs in developing deep learning models. 
With labels only for a small portion of the entire dataset, the performance approaches that of fully-supervised learning, e.g. giving competitive results in the context of image classification using less than 1\% of labeled samples~\cite{mixmatch,uda,fixmatch,noisystudent}.

Semi-supervised methods have explored different flavours of consistency or perturbation-based regularization~\cite{ssl_survey} combined with entropy minimization~\cite{entropymin} or pseudo-labeling~\cite{pseudolabel} by adding one or more extra terms to the training loss computed on the unlabeled samples.

The consistency-based regularization losses aim at exploiting the smoothness assumption~\cite{ssl_survey}, i.e. the predictions of a model should not depend on small perturbations in the input such as injecting noise~\cite{ladder,bachman14,temporalens}, adversarial perturbations~\cite{vat} or random transformations~\cite{stochastictransformations,temporalens,mixmatch,uda,fixmatch}. Pseudo-labeling and entropy minimization methods enforce instead high-confidence (low-entropy) predictions for unlabeled samples, directly exploiting the low-density assumption~\cite{ssl_survey}, i.e. decision boundaries should lie on low-density regions. These approaches are closely related to self-training~\cite{selftraining,rethinkingself,confirmationbias,noisystudent}, where pseudo-labels are extracted from a model pre-trained on labeled data and then used for training in a second step or in multiple iterative steps.

\begin{figure*}
    \centering
    \begin{tabular}{cc}
        \includegraphics[width=.33\textwidth]{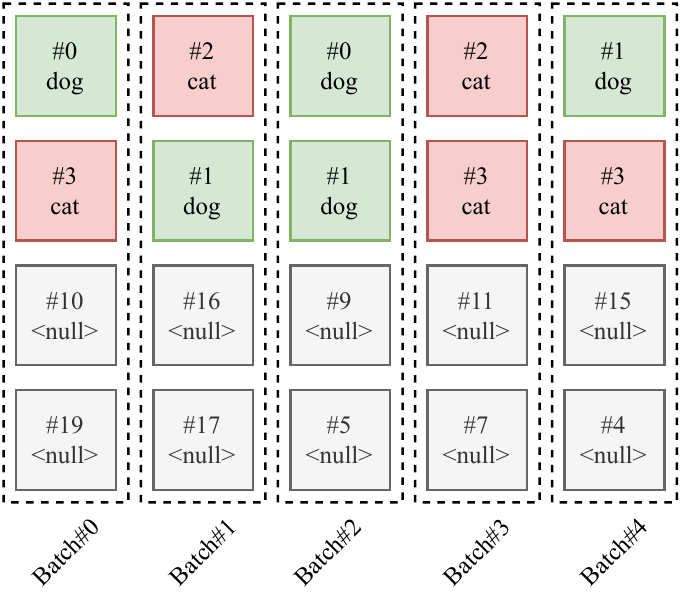}\hspace{5pt} &
        \hspace{5pt}\includegraphics[width=.33\textwidth]{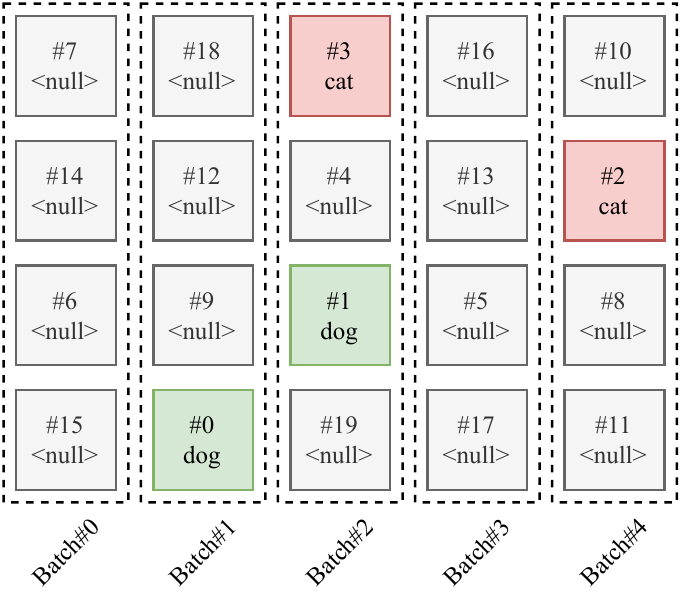} \\
        (a) Explicit setting & (b) Implicit setting
    \end{tabular}
  \caption{\small{(a) and (b) A toy dataset with 20 samples and a task with 2 classes - dog (green) and cat (red) - with 2 labeled samples per class, and a ratio of labeled samples of $20\%$. Sampled into 5 training mini-batches of 4 samples each, with over-sampling of labeled data in (a) and uniform sampling in (b). In (a) we \textit{choose} to have an equal number of labeled and unlabeled samples in each mini-batch. Due to this, each labeled sample (red/green, \#0-3) is seen 4x more often than each unlabeled sample (gray, \#4-19). In (b), each mini-batch can contain different amounts of labeled samples as it is uniformly sampled regardless of labels and each sample is seen exactly once. Labeled samples are seen 2.5x more often in the explicit setting.
  }}
\label{fig:explicitvsimplicit_batches}
\vspace{-5pt}
\end{figure*}

Yet, all semi-supervised learning methods face a common issue. Since they bootstrap their own predictions in order to use them as targets, many of these will be incorrect and can lead to so-called {\em confirmation bias}~\cite{mean,confirmationbias}, i.e. fitting incorrect labels that do not represent the true task and thus slowing down or stopping its learning. This is more likely to happen at the start of training since the predictions of the model are not much better than random and the amount of misclassified samples is large.

Most works try to alleviate this issue by using a ramp-up from zero of the weight for the semi-supervised loss term over training. This greatly reduces the influence of possibly wrong predictions on unlabeled data in the early phases of training. Thus, the only way to learn at the start is from true labels, but since these are  scarce there is little supervision available and learning is slow. This in turn makes it more difficult to get useful supervision through the semi-supervised loss.

Repeating labeled samples more often than unlabeled ones during training is a common mini-batch sampling strategy in semi-supervised learning and directly and artificially increases the influence of labeled data over learning~\cite{ladder,realistic,mixmatch, mean,vat,confirmationbias,uda,fixmatch,noisystudent}. To the best of our knowledge, this mechanism was never well studied with the goal of enhancing the training signal from true labels and it likely just happened as a side effect of common implementation choices when constructing batches from separate datasets for labeled and unlabeled samples.

In this paper, we aim at studying if it is essential to repeat labeled samples more frequently than unlabeled samples or over-sample labeled data for state-of-the-art semi-supervised methods to work well. We call this way of sampling mini-batches the \textit{explicit} semi-supervised learning setting due to its distinction between labeled and unlabeled data in how the data is presented. We compare it to an alternative setting in which all data is sampled uniformly, regardless of being labeled or not. We call the latter the \textit{implicit} semi-supervised learning setting.\footnote{In the rest of the paper we will use both nomenclatures for the sampling approaches interchangeably.} 
Figure~\ref{fig:explicitvsimplicit_batches} illustrates both approaches.

Exploring the simpler implicit setting will highlight the benefits which the common over-sampling of labeled data in the explicit setting brings and which mechanisms allow to achieve these.
Understanding the differences between both settings could then shed light on how sampling affects the learning process in semi-supervised methods and could ultimately guide future work into developing more efficient and better-performing methods.

Moreover, there are concrete reasons to avoid the explicit setting. Most importantly, trying to use it for multi-task problems easily leads into a roadblock. Samples cannot be said to be labeled or unlabeled in general, since this property can only be assigned to a sample in relation to each task. Thus, samples can be labeled for one or more tasks but unlabeled for others, labeled for no tasks or for all. This leads to different sets of samples labeled for each task, which might or might not overlap.

\begin{figure*}
    \centering
    \begin{tabular}{cc}
        \includegraphics[width=.35\textwidth]{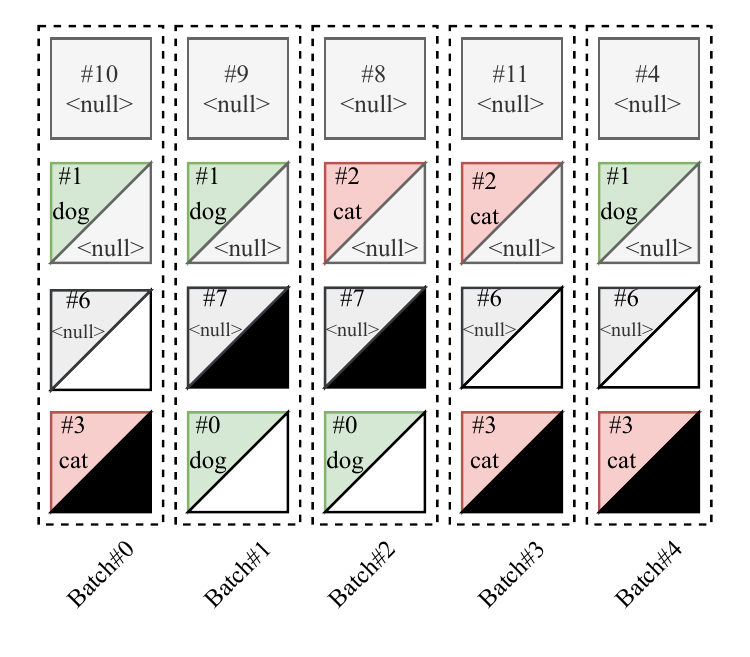} &
        \includegraphics[width=.35\textwidth]{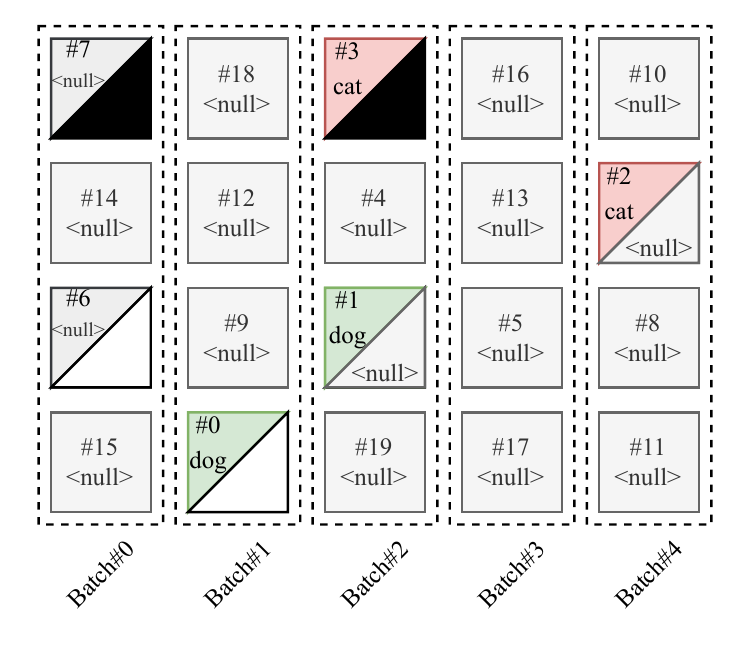} \\
        (a) Multi-task extension of explicit setting & (b) Implicit setting \\
    \end{tabular}
  \caption{\small{(a) and (b) The labeling of the dataset in Figure~\ref{fig:explicitvsimplicit_batches} is extended with a new task with 2 classes - black and white - with 2 labeled samples per class as well. In (a) we extend the over-sampling of labeled data to the multi-task case so that in each batch there is the same number of labeled and unlabeled samples for each task. In (b) the uniform sampling means the sampling is independent of the labels, just as for the single task case. 
  }}
\label{fig:mtl_explicitvsimplicit_batches}
\vspace{-5pt}
\end{figure*}

In Figure~\ref{fig:mtl_explicitvsimplicit_batches}, we compare an extension of the explicit setting for multi-task scenarios to the implicit setting. One possible direct way to recreate the explicit setting for multi-task learning is to divide each mini-batch into multiple sub-parts, one for each possible configuration of available task labels.
Such a mini-batch can include parts where all samples are labeled or unlabeled, parts with samples labeled only for one or more tasks, etc. Instead, the implicit setting can be used in multi-task problems without any modifications since the sampling is independent of the labels.

The number of sub-parts in a mini-batch constructed using this extension of the explicit setting grows exponentially with the number of tasks $T$ as $2^T$ if all label configurations are present in the dataset. Therefore, if the number of tasks is large, the number of mini-batch parts can quickly become larger than the total batch size that can be used in practice. Even with fewer tasks, the possible size of each mini-batch part becomes a new hyper-parameter which can be difficult to control since it is limited by the total batch size and the number of parts. Setting the number of tasks to one recovers the original explicit setting, with one part for labeled data for the only task in the problem and one for unlabeled data. Other sampling approaches for semi-supervised multi-task learning problems are likely possible, but our aim is to highlight the issues with the explicit setting that make it not suitable for this class of problems.

To study the impact of different approaches to sampling training mini-batches on semi-supervised performance, we compare the training characteristics of both the implicit and explicit settings. We run semi-supervised image classification experiments on CIFAR-10 and CIFAR-100~\cite{cifar}, varying the number of available labels. As a semi-supervised method we adopt FixMatch~\cite{fixmatch}, a standard, state-of-the-art method. FixMatch enforces consistency between pseudo-labels obtained from weakly transformed versions of one sample and the predictions obtained from strongly transformed versions of the same sample, incorporating in the same method pseudo-labeling and consistency regularization.
We find that both the implicit and explicit settings work for reasonably long training runs, but the latter performs better. This difference is smaller in higher-label regimes or with longer training budgets and we partially attribute it to the low efficiency of learning from true labels at the start of training when using the implicit setting, which becomes more important for shorter training runs.

In sum, this paper (i) emphasizes the importance of the choice of mini-batch sampling strategy in semi-supervised learning, (ii) analyses the differences between the common over-sampling of labeled data and standard uniform sampling throughout the training process using FixMatch, and (iii) demonstrates the viability of uniform sampling with long training budgets, enabling its use in cases where over-sampling labeled data is not possible, such as in multi-task learning.

\vspace{-5pt}
\section{Related work}
\label{sec:related}
Most semi-supervised learning methods use some mechanism that minimizes the effect of confirmation bias, either at the earlier stages of training or throughout.
One such mechanism directly ramps up the weight for the unsupervised loss term during training, starting from zero. It was originally proposed in Grandvalet and Bengio~\cite{entropymin} and is reported to be used in most works~\cite{pseudolabel,bachman14,temporalens,mean,mixmatch}. FixMatch does not use the weight ramp-up as it instead sets a threshold to use only confident predictions as pseudo-labels and averages the loss on these over the total number of unlabeled samples instead of over the number of pseudo-labels used. This achieves a similar effect to the ramp-up since it decreases the contribution of the unsupervised loss when the model produces low confidence predictions early in the training.

A second mechanism is to over-sample labeled data, i.e. repeating labeled samples more often than unlabeled ones when constructing mini-batches. This is usually implemented by splitting the labeled and unlabeled samples as two separate datasets and sampling the labeled and unlabeled parts of the mini-batch from each. Then, the first set that gets depleted is repeated as many times as needed until the other one is used up. Since the labeled dataset is typically much smaller, this results in the repetition of labeled samples at a higher rate. This approach was already used in Rasmus \textit{et al.}~\cite{ladder}\footnote{See: \href{https://github.com/CuriousAI/ladder/blob/5a8daa1760535ec4aa25c20c531e1cc31c76d911/run.py\#L74}{https://github.com/CuriousAI/ladder/blob/ 5a8daa1760535ec4aa25c20c531e1cc31c76d911/run.py\#\hspace{0pt}L74}} although without any explicit mention in the paper. Most recent works in semi-supervised learning use this setting, both using consistency-based, pseudo-labeling or self-training methods, but did not study the impact of sampling mini-batches in this way. Some methods set equal sizes for the labeled and unlabeled parts of each training mini-batch~\cite{ladder,realistic,mixmatch}, but some use a larger unlabeled batch size~\cite{mean,vat,confirmationbias,uda,fixmatch,noisystudent}.

Both FixMatch~\cite{fixmatch} and Noisy Student~\cite{noisystudent} include ablation experiments on this ratio and conclude that a larger ratio of unlabeled samples in each training mini-batch leads to better performance. Arazo \textit{et al.}~\cite{confirmationbias} is the only work we are aware of that studied the effect of keeping a minimum amount of true labels in each mini-batch in pseudo-labeling, showing that it is indeed an effective way of reducing confirmation bias. Only Temporal Ensembling~\cite{temporalens} and one of the experiments in Mean Teacher~\cite{mean} use the implicit setting. However, the latter mentions that repeating labeled samples can be beneficial since "the supervised training signal is strong enough early on to train quickly and prevent getting stuck into uncertainty"~\cite{meanteacher_repo}. All methods relying on the explicit setting can technically be adapted to use the implicit setting instead.

\section{Explicit vs. implicit setting}
\label{sec:experiments1}
\label{sec:experiments:fixmatch}
We aim at understanding how essential the explicit setting
is for good performance in semi-supervised learning, and whether
the implicit one
can achieve competitive results. To do so, we will compare the final classification performance of both settings.

\subsection{Experimental setting}
We compare the explicit and implicit settings head to head when using FixMatch~\cite{fixmatch}, which uses asymmetric perturbations in the input space in a siamese network configuration with two branches. The teacher branch gets weakly augmented samples (only flip-and-shift operations) and generates pseudo-labels from them, filtering out those below a certain confidence threshold. The student branch gets strongly augmented samples via RandAugment~\cite{randaug} and is trained using the standard cross-entropy loss against both true labels (if available) and the pseudo-labels generated by the teacher.

We use the semi-supervised image classification problem on the CIFAR-10~\cite{cifar} dataset for the main experiments, as it is a standard benchmark in the deep semi-supervised learning literature~\cite{realistic}. We use three different amounts of labeled samples - 40, 250 and 4000 - simulating respectively a problem with extremely limited labeled data at only four labels per class, a scarcely-labeled problem with approximately 0.5\% labeled samples, and a semi-supervised learning problem with a larger amount of labeled samples with approximately 9\% labeled samples. We reserve 10\% of the total samples in the training set for validation and use the rest as training data, which we split into labeled and unlabeled subsets.
All splits are done preserving the class balance. The validation set is used for early stopping and for hyper-parameter tuning based on random search with a budget of 100 trials per experiment. We pick the hyper-parameters leading to the highest validation accuracy on a single seed and data split. We tune the following hyper-parameters for each case: learning rate, batch size, weight of the semi-supervised loss, and the ratio of labeled data in each batch for the explicit setting,. Finally, we report the final classification accuracy values on the test set as the mean and standard deviation over five different random seeds and five different splits of the labeled samples in the training set.

In addition, we perform experiments on CIFAR-100 with 2500 and 10000 labels, following the same setting as for CIFAR-10 but not tuning the hyper-parameters. Instead, we use the ones reported in FixMatch~\cite{fixmatch} for the explicit setting and our best single guess of a decent set of hyper-parameters for the implicit setting.

For CIFAR-10 experiments, we use a Wide ResNet-28-2~\cite{wideresnet} (WRN-28-2) as the base network with a linear classifier after the pooling layer and train it via stochastic gradient descent with Nesterov momentum of $0.9$~\cite{momentum,nesterov,importance_init_momentum}, a cosine learning rate decay schedule~\cite{sgdr}, and weight decay of $0.0005$. For CIFAR-100, we use WRN-28-5 and a smaller weight decay of $0.0001$. Evaluation is done on an exponential moving average of the model parameters with decay $0.999$~\cite{fixmatch}, which we found gave more stable results and significantly better performance early in the training.

Each training run has a maximum budget for the total number of samples that each model sees during training. We set this budget to be equivalent to $1000$ epochs on the full training set, i.e. $45$ million samples, after removing 5000 samples from the training set for validation. Therefore we refer to an epoch as one pass through as many samples as the size of the training set, regardless of whether each sample has an associated label or not, or how many times it has been repeated.

Other works~\cite{realistic,fixmatch} have used a fixed number of labeled samples or a fixed number of training steps instead. We believe this approach leads to unfair comparisons between supervised baselines or when using different label ratios in each batch since for higher ratios the models will effectively see many more samples over training. Moreover, comparing the total amount of samples seen aligns better with the amount of compute used for each training run. Using this definition of training length, MixMatch~\cite{mixmatch} training runs go through $\sim134$ million samples since they use a total batch size of 128 (64 for labeled samples and 64 for unlabeled samples) but they only count the labeled ones towards their definition of total length. FixMatch, using an unlabeled batch size seven times larger, uses a total batch size of $512$ and the total number of samples seen during training is $\sim536$ million.

We choose to use a smaller budget on purpose to better reflect what would happen in practical applications and to keep contained the computational budget required for our experiments. We noticed that for smaller amounts of labeled data longer training budgets are important, so we include longer experiments in such cases as well.

\subsection{Experimental results}

\begin{table}[t]
\centering
\begin{adjustbox}{max width=\columnwidth}
\begin{tabular}{c c c c c}
\toprule
    \textbf{CIFAR-10} & 40 labels & 250 labels & 4000 labels & All labels\\ 
    \midrule
    Supervised  & $36.77\pm4.48$ & $59.88\pm.73$       &      $87.39\pm.20$  &  $96.54\pm.11$     \\
    \midrule
    FixMatch(E) & $83.44\pm6.68$  & $93.01\pm.58$       &      $94.89\pm.16$ &  -  \\ 
    FixMatch(I)  &   $73.90\pm8.04$ & $91.42\pm.98$       &      $94.84\pm.05$&   -  \\
    \midrule
    FixMatch(E)~6x & $85.40\pm2.83$ & $94.40\pm.78$       &      - &  -  \\ 
    FixMatch(I)~6x  &   $86.59\pm4.14$ & $94.00\pm.54$       &      - &   -  \\
    \midrule
    \textbf{CIFAR-100} & & 2500 labels & 10000 labels & All labels\\ 
    \midrule
     Supervised & & $47.26$ & $67.58$ & $82.38$ \\
     \midrule
     FixMatch(E) & & $62.86$ & $74.61$ & - \\
     FixMatch(I) & & $55.53$ & $71.83$ & - \\
\bottomrule\\
\end{tabular}
\end{adjustbox}

\caption{\small{Classification accuracy on test data for supervised baselines and FixMatch with both implicit (I) and explicit (E) settings using different amounts of labels. Shown as mean and standard deviation on 5 runs with different seeds and data splits for each CIFAR-10 experiment and a single seed and data split for CIFAR-100. 6x indicates six times longer training budget.}}
\label{table:results}
\end{table}

Table~\ref{table:results} shows the classification accuracy of our models trained on CIFAR-10 and CIFAR-100 with different amounts of labels when using FixMatch in either the explicit or implicit settings.

We observe that both settings outperform supervised baselines by a significant margin in CIFAR-10. For 4000 labels both settings perform similarly, however we see a larger gap favouring the explicit setting by $1.6$ percentage points when only using 250 labels. A significantly larger gap of $9.5$ percentage points appears when using 40 labels, although both settings show a much larger standard deviation. Comparing individual runs on each data split for both settings, we can see that the large standard deviations are caused mainly by the choice of labeled samples in each data split but also that the implicit setting is more sensitive to it.

We hypothesize that this gap is due to the much smaller amount of labeled samples present (on average) in each training mini-batch for the lower-label regimes in the implicit setting, where for example around only one labeled sample appears for every 200 unlabeled samples for the 250-label case. Due to this lack of direct supervision at the start, the model learns at a very slow pace. This is because the model is initially relying only on labeled data as the predictions on unlabeled data are below the confidence threshold. A longer training budget of 6000 epochs improves the performance of both settings. The gap in test accuracy between them reduces over training, and both settings end up performing on par.

We see similar results for CIFAR-100. The gap between settings is larger at lower label regimes, reaching $7.3$ percentage points better accuracy for the explicit setting with 2500 labels and $2.8$ percentage points with 10000 labels, and both perform better than the supervised baselines.\\

\begin{figure*}
    \centering
    \begin{tabular}{cccc}
        \hspace{-5pt}\includegraphics[width=.49\columnwidth]{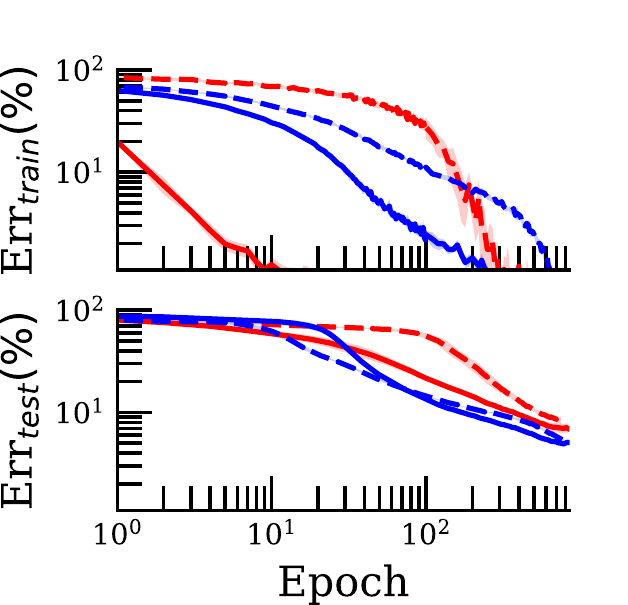} &
        \hspace{-5pt}\includegraphics[width=.49\columnwidth]{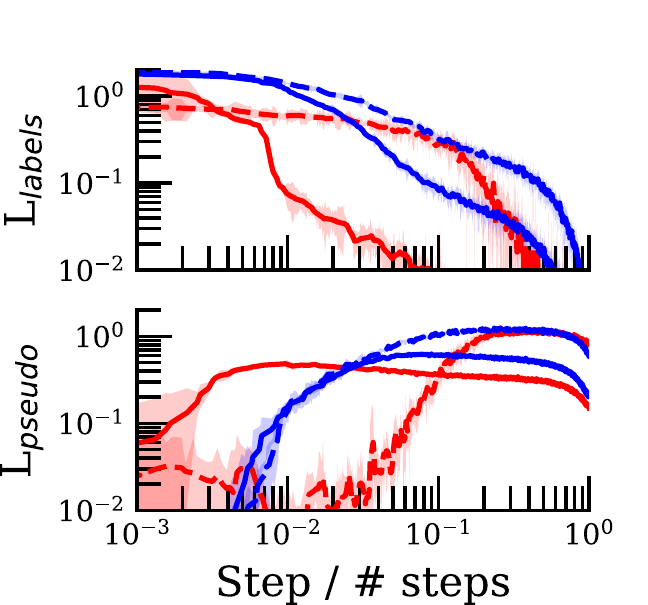} &
        \hspace{-5pt}\includegraphics[width=.49\columnwidth]{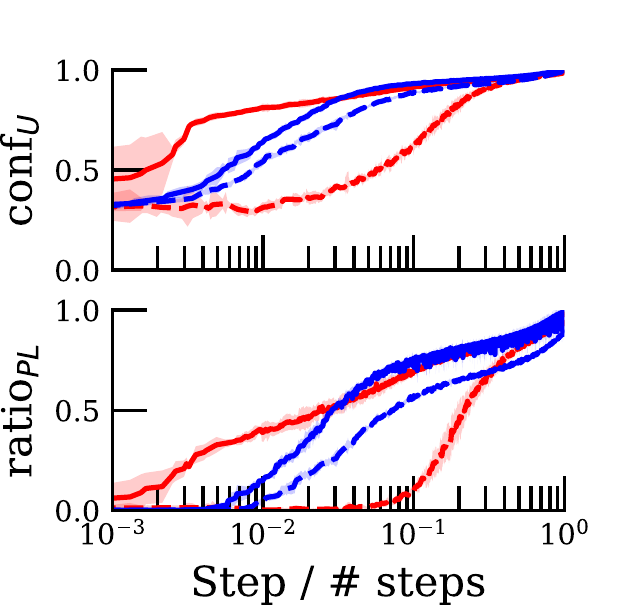} &
        \hspace{-5pt}\includegraphics[width=.49\columnwidth]{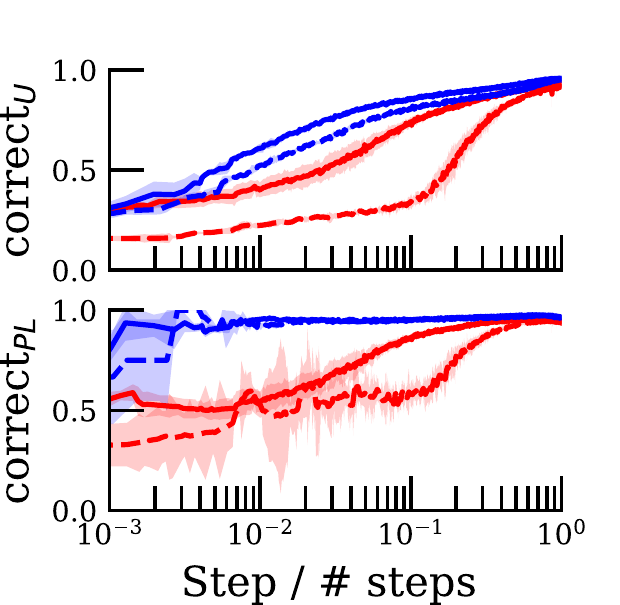} \\
        \small{(a)} & \small{(b)} &
         \small{(c)} & \small{(d)}\\
    \end{tabular}
   \caption{\small{Training curves for CIFAR-10 with 250 labels (red) or 4000 labels (blue) using FixMatch in the explicit (solid line) or implicit (dashed) settings. (a) Top: training error on labeled samples. Bottom: test error on the unseen test set. Values at the end of each epoch. (b) Top: evolution over training of cross-entropy loss on labeled training data. Bottom: semi-supervised cross-entropy loss between pseudo-labels and predictions on unlabeled data. (c) Top: average confidence of predictions on unlabeled samples in each batch. Bottom: ratio of predictions used as pseudo-labels. (d) Top: Ratio of correct predictions on unlabeled data. Bottom: Ratio of correct pseudo-labels on each batch.}
   }
\label{fig:explicitvsimplicit}
\end{figure*}

\section{Analysis}

\textbf{Training dynamics} We analyze in Figure~\ref{fig:explicitvsimplicit} the training curves of models in both settings for 250 and 4000 labels to illustrate the impact of their different training dynamics.
We can see that the explicit setting does indeed fit quicker the labeled data; training error, shown in (a, top), and supervised loss, in (b, top), both decrease quickly thanks to more direct supervision from repeating labeled samples. The models start giving confident predictions earlier, as shown in (c, top), and thus the number of pseudo-labels, in (c, bottom), and the semi-supervised loss, in (b, bottom), grow quicker. On the other hand, for the implicit setting, the model takes much longer to fit the training data and to produce high-confidence predictions that will be used as pseudo-labels. Therefore, the semi-supervised loss grows much slower, having no effect for the first 100 epochs while learning is very slow and test accuracy is barely better than random. This clearly points to a lack of direct supervision from true labels early in training as one cause for the implicit setting lagging behind in terms of final test error with our relatively short training budget with fewer labeled samples.

Moreover, we make use of privileged information (the labels dropped to simulate unlabeled samples) to compute and show in Figure~\ref{fig:explicitvsimplicit}~(d) the ratio of correct predictions on unlabeled samples and the ratio of the correct pseudo-labels, i.e. the subset of predictions on unlabeled data whose confidence was above the threshold of $0.95$. We see that not only are pseudo-labels used earlier in the explicit setting but are also cleaner, i.e. more of them are correct, from early on and can thus help guide learning of the true classification task better. Also, we can see that the usage of confident predictions on unlabeled data as pseudo-labels is effective since the subset of chosen pseudo-labels has a ratio of correctness consistently higher than that of all predictions on unlabeled data.

This analysis holds for both amounts of labeled samples shown, although the differences are much smaller for the 4000-label experiments. In this last case, the implicit setting has a lower test error earlier during training which we hypothesize might be due to the larger over-fitting to labeled samples in the explicit case, before the loss on pseudo-labels starts having a regularizing effect and the model can generalize to unseen data.

\textbf{Training without direct supervision }
We hypothesize that supervision from true labels loses importance when the model can already generate cleaner pseudo-labels, which could explain why either using more labels or longer training budgets reduces the performance gap between settings.

To verify this hypothesis, we run FixMatch with the semi-supervised loss only but starting training from different supervised checkpoints. All models improve their test accuracy at first - the best goes from $87.39\%$ to $92.9\%$ - but their accuracy ends up dropping suddenly to levels close to random while fitting the pseudo-labels. This can be attributed to the accumulation of confirmation errors and the lack of supervision from true labels to correct for these. We see this happen as well for long experiments with the implicit setting and few labels, but with learning recovering after the dip. Thus, we conclude that seeing some labeled samples is still required to guide learning to the true task even when few pseudo-labels are incorrect.

\section{Conclusion}

\label{sec:conclusion}
Our experiments in semi-supervised image classification show that using the implicit setting incurs a penalty in final accuracy compared to using the explicit setting. How large this gap becomes depends, however, on the amount of labels and training budget. Both settings still outperform supervised baselines by a large margin. 
The explicit setting is significantly more efficient in the beginning but the difference between both settings is reduced over training and both perform on par when more labeled data is available and for longer training runs. However, seeing labeled samples remains important even in the later stages of training, as it avoids falling into confirmation bias. The more general aspects like the required amount of labels and the most effective way to use them are yet to be studied.
Overall, these are important results since the implicit setting is much simpler to use, potentially allowing semi-supervised learning on settings like partially-labeled multi-task problems which would not be possible in the conventional explicit setting.

This analysis was done in the context of FixMatch, which we believe is a good representative of modern semi-supervised deep learning methods combining consistency regularization and pseudo-labeling approaches. Future research will be directed to see if the results also apply to semi-supervised methods that are substantially different.

Finally, we suggest two possible alternative approaches to mini-batch sampling in the multi-task scenario. First, if there is a large enough amount of samples labeled for all tasks, a version of the explicit setting could be constructed using \textit{fully} labeled samples versus \textit{partially} labeled samples. Second, we could construct a variation of the implicit setting with a guaranteed minimum number of samples labeled for each task so that there is always \textit{some} direct supervision from true labels for each task, in each mini-batch. Studying these and other possible sampling methods will be an interesting direction for future work.

\section*{Acknowledgments}
This work was partially supported by the Wallenberg AI, Autonomous Systems and Software Program (WASP) funded by the Knut and Alice Wallenberg Foundation.

The support by the Swedish Research Council through grant agreement no. 2016-04022 is also gratefully acknowledged.

The computations were partially enabled by resources provided by the Swedish National Infrastructure for Computing (SNIC) at Chalmers Centre for Computational Science and Engineering (C3SE) partially funded by the Swedish Research Council through grant agreement no. 2018-05973.


\bibliography{egbib}

\end{document}